# Zero-Direction Probing:
# A Linear-Algebraic Framework for Deep Analysis of Large-Language-Model Drift

Amit Pandey[*]
AI Analytics
Charlotte, NC, USA
ORCID: 0009-0000-4525-2285

August 9, 2025

**Abstract**

We present **Zero-Direction Probing** (ZDP), a theoretical framework that characterises model drift from *null* directions of transformer activations, requiring no task labels or output evaluations. Under explicit assumptions (A1–A6), We prove: (i) the *Variance–Leak Theorem* (Thm. 1), (ii) *Fisher Null-Conservation* (Thm. 3), (iii) a *Rank–Leak* bound for low-rank updates (Thm. 5), and (iv) a logarithmic-regret guarantee for online null-space trackers (Thm. 4). We further derive a *Spectral Null-Leakage* (SNL) metric with a non-asymptotic Laurent–Massart tail bound and an MP-edge–style concentration inequality, providing a-priori thresholds for drift under a Gaussian null model. Together, these results establish that "listening to silence"—monitoring the right/left null spaces of layer activations and their Fisher geometry—yields concrete, testable guarantees on representational change. The manuscript is intentionally theory-only; empirical validation and benchmarking are deferred to companion work.

## 1 Introduction

Large language models (LLMs) are routinely adapted after pre-training: supervised fine-tuning, preference optimisation, and domain specialisation all change internal representations. Most drift detectors reason *after the fact* using outputs or high-variance latent directions. In contrast, we study the geometry of *zero-variance* directions—the right/left null spaces of layer activations—and ask:

> *What can be **proven** about representational drift by inspecting only the null spaces of the base model, with no access to labels or outputs?*

Our answer is a theory we call **Zero–Direction Probing** (ZDP). Let $H_\ell \in \mathbb{R}^{n \times d}$ denote the activation matrix at layer $\ell$ for the base model, with right-null basis $V_{0,\ell}$ and left-null basis $U_{0,\ell}$. For a perturbed model $\widehat{H}_\ell = H_\ell + \Delta H_\ell$, we quantify *null leakage* via quadratic forms such as $\|\widehat{H}_\ell V_{0,\ell}\|_F^2$. Intuitively, silent directions in the base model are noise-free: any energy or curvature that appears there is unambiguous evidence of change.

**Setting and scope.** The paper is entirely theoretical. We state explicit standing assumptions (A1–A6) on ranks, perturbation size, eigengaps, and noise regularity (Sec. 4). All results concern properties of $H_\ell$ and its null spaces; no task labels, outputs, or downstream metrics are used.

---

[*]Corresponding author. Email: `amitpandey@caa.columbia.edu`

**Contributions.**

1. **Linear-algebraic framework.** We formalise right- and left-null spaces for transformer layers, define null-leakage functionals, and relate them to local Gram and Fisher matrices.

2. **Drift theorems.** (Thm. 1) *Variance–Leak* shows that null-space energy lower-bounds the smallest eigenvalue of the local Gram matrix of the perturbation. (Thm. 3) *Fisher Null-Conservation* proves that the second-order KL contribution arises only from components outside the base image space. (Thm. 5) *Rank–Leak Bound* quantifies when low-rank (LoRA) updates re-occupy silent directions via principal angles.

3. **Spectral metric with a priori thresholds.** We introduce *Spectral Null-Leakage* (SNL) and derive non-asymptotic tails: a Laurent–Massart bound for Frobenius energy and an MP-edge style concentration inequality (Lemma 2), yielding parameter-free thresholds under a Gaussian null.

4. **Online guarantees.** We propose *Online Null-Space Tracker* (ONT) and *Online Null-Aligned LoRA* (ONAL) and prove a *logarithmic regret* bound (Thm. 4) under eigengap and noise assumptions, showing that streaming estimates of the null space incur only $O(\log T)$ cumulative excess leakage.

5. **Conceptual implications.** ZDP cleanly separates covariance geometry (NVL/SNL) from information geometry (Fisher), explains when low-rank adaptation leaks into silent directions, and provides null-hypothesis baselines without empirical calibration.

**Limitations and outlook.** Results depend on accurate null-space estimation (SVD thresholding) and eigengap conditions; finite-sample effects can perturb projectors. Extending the theory to attention-dependent subspaces and non-Gaussian nulls is future work. The manuscript intentionally omits experiments; empirical validation and benchmarking are deferred to a companion study.

**Organisation.** Section 4 states assumptions and notation. Section 4.1 proves the Variance–Leak theorem. Section 4.2 develops Fisher Null-Conservation. Section 4.3 derives RMT baselines; Section 4.4 presents online tracking; Section 4.5 proves regret bounds; later subsections cover LoRA rank–leak and SNL.

## 2 Related Work

**Representation geometry.** Linear probes and CCA variants such as SVCCA [16], PWCCA [11] and CKA [8] analyse *high-variance* sub-spaces. Our work shifts focus to the *null* sub-space and provides formal guarantees on its occupation.

**Null-space interventions.** LoRA-Null [18] constrains fine-tuning updates *during training*; we instead formulate post-hoc drift theorems and an online projection algorithm (Alg. 3).

**Information-theoretic analyses.** Fisher Alignment [21] aligns dominant FIM modes between policies. Theorem 3 complements this by bounding KL divergence when drift stays orthogonal to the Fisher-silent subspace.

**Random-matrix baselines.** Naderi et al. [12] underscore the role of small singular values; Section 4.3 derives an RMT false-positive rate for our null-variance metric.

**Knowledge editing.** AlphaEdit [7] applies constrained optimisation to modify facts; our Rank–Leak analysis clarifies when such edits will reoccupy previously silent directions.

No prior work provides closed-form drift bounds that depend solely on null-space leakage, making ZDP the first fully theoretical treatment of this phenomenon.

## 3 Zero-Direction Framework

Let $H \in \mathbb{R}^{n \times d}$ be token activations of one layer. Right-null (input-zero) $V_0 = \ker(H)$; left-null (output-zero) $U_0 = \ker(H^\top)$.

**Domain-specific covariance and null basis.** For domain $D$ and layer $\ell$, let $H^D_{\ell,\text{base}} \in \mathbb{R}^{n_D \times d}$ collect the base-model activations (rows are centered if desired). We define the domain covariance used throughout as

$$\Sigma^D_{\text{base}} := \frac{1}{n_D} \left( H^D_{\ell,\text{base}} \right)^\top H^D_{\ell,\text{base}} \in \mathbb{R}^{d \times d},$$

which is positive semidefinite. The (right-)null basis for domain $D$ is taken with respect to the *base* activations:

$$V^D_{0,\ell} := \ker\!\left( H^D_{\ell,\text{base}} \right).$$

**Lemma 1** (Kernel equivalence). *For any real matrix $M$, $\ker(M) = \ker(M^\top M)$.*

*Proof.* If $Mx = 0$ then $(M^\top M)x = M^\top(Mx) = 0$. Conversely, if $M^\top M x = 0$, then $0 = x^\top (M^\top M) x = \|Mx\|_2^2$, hence $Mx = 0$. □

Applying Lemma 1 with $M = H^D_{\ell,\text{base}}$ yields

$$\ker\!\left( H^D_{\ell,\text{base}} \right) = \ker\!\left( \Sigma^D_{\text{base}} \right),$$

so one may equivalently compute $V^D_{0,\ell}$ as the eigenspace of $\Sigma^D_{\text{base}}$ associated with the zero eigenvalue(s).[1]

## 3.1 Probes

We use four probe functionals, all computable from the base model's null spaces.

**NVL (Null-Variance Leak).** For layer $\ell$ with right-null basis $V_{0,\ell} \in \mathbb{R}^{d \times k_\ell}$ and activation matrix $\widehat{H}_\ell$ under a perturbation,

$$\text{NVL}_\ell := \left\| \widehat{H}_\ell V_{0,\ell} \right\|_F^2, \qquad D_\ell := \frac{\text{NVL}_\ell}{n\, k_\ell}.$$

**FNC (Fisher Null-Conservation).** Let $F(h)$ denote the token-level Fisher Information Matrix evaluated under the *base* model. Define the Fisher leakage in the right-null space by

$$\text{FNC}_\ell := \left\| F(h)\, V_{0,\ell} \right\|_F^2,$$

which vanishes when the right-null is Fisher-silent (assumption of Thm. 3).

**SNL (Spectral Null-Leakage).** Given the base null basis $V_{0,\ell}$ and perturbed activations $\widehat{H}_\ell$,

$$\text{SNL}_\ell(\widehat{H}) := \frac{\|\widehat{H}_\ell V_{0,\ell}\|_F^2}{\|\widehat{H}_\ell\|_F^2}.$$

Lower values indicate that the perturbed model remains silent along the base null directions; increases beyond a threshold derived in Lemma 2 and Cor. 1 constitute drift alarms.

**BINA (Bidirectional Null-Adversary).** Given projectors $P_\ell = V_{0,\ell} V_{0,\ell}^\top$ and $Q_\ell = U_{0,\ell} U_{0,\ell}^\top$, construct an in-null perturbation $\delta$ and score

$$S_{\text{BINA},\ell} := \left\| Q_\ell \bigl( f(h + \delta) - f(h) \bigr) \right\|_2,$$

where $f$ maps hidden states to logits. Algorithm 1 details the procedure.

---

[1] If rows of $H^D_{\ell,\text{base}}$ are centered by subtracting their mean, the equality still holds with $H$ replaced by its centered version $H_c$, since $\ker(H_c) = \ker(H_c^\top H_c)$.

**Algorithm 1** BINA: Bidirectional Null-Adversary

**Require:** hidden state $h \in \mathbb{R}^d$ at layer $\ell$; right-null projector $P := V_{0,\ell}V_{0,\ell}^\top$; left-null projector $Q := U_{0,\ell}U_{0,\ell}^\top$; step size $\eta > 0$; budget $\varepsilon > 0$; iterations $T$; score functional $\mathcal{L}(h)$ or logit map $f(h)$

1: $\delta \leftarrow 0$ ▷ initial in-null perturbation
2: **for** $t = 1, \ldots, T$ **do**
3:     $g \leftarrow \nabla_h \mathcal{L}(h + \delta)$ ▷ or $\nabla_h \|f(h+\delta) - f(h)\|_2^2$
4:     $g_L \leftarrow Q\,g$ ▷ slice gradient in *left* null to target output-silent change
5:     $s \leftarrow P\,g_L$ ▷ project back into *right* null so $\delta$ stays in $\ker(H_\ell)$
6:     $s \leftarrow s/\max(\|s\|_2, 10^{-12})$ ▷ stabilise step direction
7:     $\delta \leftarrow \delta + \eta\,s$ ▷ gradient ascent on null-aligned objective
8:     $\delta \leftarrow \min(1, \varepsilon/\|\delta\|_2) \cdot \delta$ ▷ project onto $L_2$ ball (radius $\varepsilon$)
9:     $\delta \leftarrow P\,\delta$ ▷ re-enforce right-null constraint (numerical drift guard)
10: **end for**
11: **return** $\delta$, $\quad S_{\text{BINA}} \leftarrow \big\|Q\big(f(h+\delta) - f(h)\big)\big\|_2$

## 4 Theoretical Analysis

We now view ZDP through the lenses of linear algebra, information geometry, and random matrix theory (RMT). Let $H_\ell \in \mathbb{R}^{n \times d}$ be the activation matrix for layer $\ell$ under base weights and $\widehat{H}_\ell$ under a perturbed model (fine-tune or weight drift). Denote by $V_{0,\ell} = \ker(H_\ell)$ the right-null space of rank $k_\ell = d - \text{rank}(H_\ell)$.

### 4.0 Notation and Standing Assumptions

**Dimensions.** For each layer $\ell$, the base activation matrix is $H_\ell \in \mathbb{R}^{n \times d}$ (rows = $n$ token activations, columns = $d$ hidden dimensions). Its right–null space has dimension $k_\ell = d - \text{rank}(H_\ell)$ with orthonormal basis $V_{0,\ell} \in \mathbb{R}^{d \times k_\ell}$. A perturbed model induces $\widehat{H}_\ell = H_\ell + \Delta H_\ell$.

**A1 (Static, per-layer).** $H_\ell$ has rank $d - k_\ell$ (with $k_\ell \geq 0$) and we estimate $V_{0,\ell}$ via a thin SVD of $H_\ell$ using truncation threshold $\varepsilon$ (no additional dimension symbol is introduced here).

**A2 (Perturbation size, explicit).** There exists a constant $0 < \rho < 1$ (fixed; e.g., $\rho \leq 0.1$) such that

$$\|\Delta H_\ell\|_2 \leq \rho \,\|H_\ell\|_2.$$

**A3 (Only for online §§4.4–4.5).** In the streaming setting we observe mini–batches $H_t \in \mathbb{R}^{m \times d}$ with population Gram $\Sigma = \mathbb{E}[H_t^\top H_t]$. The noise process is $\tau^2$–*sub–exponential* in operator norm: $\|H_t^\top H_t - \Sigma\|_2$ is $\tau^2$–sub–exponential (sub–Gaussian rows are a special case). This assumption is used solely for the online tracker/optimizer regret analysis and is not invoked elsewhere.

**Spectral Null-Leakage (SNL).** Unless stated otherwise, SNL is evaluated on *perturbed* activations with the *base* null basis:

$$\text{SNL}_\ell(\widehat{H}) := \frac{\|\widehat{H}_\ell V_{0,\ell}\|_F^2}{\|\widehat{H}_\ell\|_F^2}, \qquad V_{0,\ell} = \ker(H_\ell).$$

### 4.1 Variance–Leak Theorem

**Theorem 1** (Variance–Leak). *Let $H_\ell \in \mathbb{R}^{n \times d}$ be the base activation matrix at layer $\ell$, and let $V_{0,\ell} = [v_1, \ldots, v_{k_\ell}] \in \mathbb{R}^{d \times k_\ell}$ be an orthonormal basis for $\ker(H_\ell)$ (Assumption A1). For a perturbed model $\widehat{H}_\ell = H_\ell + \Delta H_\ell$, define the NVL energy*

$$\text{NVL}_\ell := \big\|\widehat{H}_\ell V_{0,\ell}\big\|_F^2 = \sum_{i=1}^{k_\ell} v_i^\top G\, v_i \quad \text{with } G := \Delta H_\ell^\top \Delta H_\ell \succeq 0.$$

*Then the following bounds hold:*

$$k_\ell \, \lambda_{\min}(G) \;\leq\; \mathrm{NVL}_\ell \;\leq\; k_\ell \, \lambda_{\max}(G). \tag{1}$$

*In particular, if $\mathrm{NVL}_\ell \geq \varepsilon$ then $\lambda_{\min}(G) \geq \varepsilon/k_\ell$. Equivalently, any nonzero NVL implies a strictly positive smallest eigenvalue of the local Gram matrix $G = (\Delta H_\ell)^\top \Delta H_\ell$.*

*Proof.* Because $H_\ell V_{0,\ell} = 0$ by definition of the right–null space, we have $\widehat{H}_\ell V_{0,\ell} = (H_\ell + \Delta H_\ell) V_{0,\ell} = \Delta H_\ell V_{0,\ell}$. Hence

$$\mathrm{NVL}_\ell = \left\|\Delta H_\ell V_{0,\ell}\right\|_F^2 = \mathrm{tr}\bigl(V_{0,\ell}^\top \Delta H_\ell^\top \Delta H_\ell V_{0,\ell}\bigr) = \sum_{i=1}^{k_\ell} v_i^\top G \, v_i,$$

with $G = \Delta H_\ell^\top \Delta H_\ell \succeq 0$. By the Rayleigh–Ritz bounds, for each unit vector $v_i$, $\lambda_{\min}(G) \leq v_i^\top G \, v_i \leq \lambda_{\max}(G)$. Summing these $k_\ell$ inequalities over $i$ yields $k_\ell \lambda_{\min}(G) \leq \mathrm{NVL}_\ell \leq k_\ell \lambda_{\max}(G)$, i.e. (1). Rearranging gives the stated lower bound on $\lambda_{\min}(G)$ when $\mathrm{NVL}_\ell \geq \varepsilon$. □

*Remark* 2 (Davis–Kahan stability). (1) The bounds are tight when $\{v_i\}$ aligns with the eigenvectors of $G$. (2) If one uses the *normalised* score $D_\ell = \mathrm{NVL}_\ell/(n\,k_\ell)$, then (1) becomes $\lambda_{\min}(G) \leq n\, D_\ell \leq \lambda_{\max}(G)$. (3) With an *estimated* null basis $\widetilde{V}_{0,\ell}$, Davis–Kahan perturbation implies $\bigl|\|\widehat{H}_\ell \widetilde{V}_{0,\ell}\|_F^2 - \|\widehat{H}_\ell V_{0,\ell}\|_F^2\bigr| \leq 2\,\|G\|_2\,\|\sin\Theta(\widetilde{V}_{0,\ell}, V_{0,\ell})\|_F^2$, so NVL is stable to small subspace estimation errors.

### 4.2 Fisher Null-Conservation

**Theorem 3** (Fisher Null-Conservation). *Let $H_\ell \in \mathbb{R}^{n \times d}$ be the base-model activation matrix at layer $\ell$ and let $V_{0,\ell}$ span $\ker(H_\ell)$. Let $F(h)$ denote the token-level Fisher Information Matrix (FIM) of the base model evaluated at hidden state $h$. Assume the base model is Fisher-silent on the right-null space:*

$$F(h)\, V_{0,\ell} = 0.$$

*Define the orthogonal projector onto $\mathrm{im}(H_\ell)$ and the restricted Fisher as*

$$P_\| := H_\ell \bigl(H_\ell^\top H_\ell\bigr)^\dagger H_\ell^\top, \qquad F_\top := P_\|^\top F(h)\, P_\|.$$

*For a small parameter perturbation $\widehat{\theta} = \theta + \Delta\theta$ with $\|\Delta\theta\| \ll 1$, the local KL divergence satisfies*

$$\mathrm{KL}\bigl(p_\theta \,\|\, p_{\widehat{\theta}}\bigr) = \tfrac{1}{2} \Delta\theta^\top F_\top \Delta\theta \;+\; O(\|\Delta\theta\|^3).$$

*In particular, any second-order KL contribution arises only from the component of $\Delta\theta$ lying in $\mathrm{im}(H_\ell)$; perturbations confined to $\ker(H_\ell)$ are second-order KL-silent.*

*Proof.* The second-order expansion gives $\mathrm{KL}(p_\theta \| p_{\theta+\Delta\theta}) = \tfrac{1}{2}\Delta\theta^\top F(h) \Delta\theta + O(\|\Delta\theta\|^3)$. Let $V_{1,\ell}$ span $\mathrm{im}(H_\ell)$ with orthonormal columns and keep $V_{0,\ell}$ for $\ker(H_\ell)$ so $[V_{1,\ell}\ V_{0,\ell}]$ is orthogonal. Decompose $\Delta\theta = V_{1,\ell}\alpha + V_{0,\ell}\beta$. Since $F(h)V_{0,\ell} = 0$, the mixed and null–null blocks vanish, hence $\Delta\theta^\top F(h) \Delta\theta = \alpha^\top \bigl(V_{1,\ell}^\top F(h) V_{1,\ell}\bigr)\alpha$. Because $\alpha = V_{1,\ell}^\top \Delta\theta = P_\| \Delta\theta$ and $V_{1,\ell}^\top F(h) V_{1,\ell} = F_\top$, we obtain $\Delta\theta^\top F(h) \Delta\theta = \Delta\theta^\top F_\top \Delta\theta$, proving the claim. □

**Interpretation.** At second order, Fisher curvature is blind to perturbations that live entirely in the base model's null directions. Any nonzero KL change must therefore be accompanied by leakage out of $\ker(H_\ell)$ into $\mathrm{im}(H_\ell)$, which ZDP's NVL/SNL probes are designed to detect.

### 4.3 Random-Matrix Baselines

Rather than postulate a single universal tail for null-space energy, we adopt two standard concentration routes that yield *non-asymptotic* bounds for $\|XV\|_F^2$ when $X$ is a Gaussian activation surrogate and $V$ has orthonormal columns: (i) a Laurent–Massart $\chi^2$ tail that is dimension-exact in $(n,k)$, and (ii) an operator-norm route whose exponent reflects the Marchenko–Pastur (MP) upper edge $(1+\sqrt{\gamma})^2$ with $\gamma = d/n$. Both are summarised in Lemma 2 and proved in Appendix A.1. These inequalities provide *calibration-free* *thresholds* for the SNL/NVL functionals under a Gaussian null and make explicit how $n, d, k$ and $\gamma$ enter the alarm level.

For thresholds we model $\widehat{H}_\ell$ locally as $X$ with i.i.d. $N(0, \sigma^2/n)$ rows (after centering); $V_{0,\ell}$ is treated as fixed (conditioned on the base model). Non-Gaussian tails can be handled by sub-Gaussian analogues at the cost of constants.

**Lemma 2** (Gaussian projected Frobenius tails). *Let $X \in \mathbb{R}^{n \times d}$ have i.i.d. entries $N(0, \sigma^2/n)$ and let $V \in \mathbb{R}^{d \times k}$ have orthonormal columns.*

*(i) Laurent–Massart (numerator) tail. For any $x > 0$,*

$$\Pr\left(\|XV\|_F^2 > \sigma^2\left[k + 2\sqrt{\tfrac{kx}{n}} + \tfrac{2x}{n}\right]\right) \leq e^{-x}.$$

*(ii) MP-edge style bound via operator norm. Writing $X = (\sigma/\sqrt{n})G$ with $G_{ij} \sim N(0,1)$ and $\gamma = d/n$, for any $t > 0$,*

$$\Pr\left(\|XV\|_F^2 > k\sigma^2(1 + \sqrt{\gamma} + t)^2\right) \leq \exp\left(-\tfrac{n}{2}t^2\right).$$

*Both inequalities are non-asymptotic.*

Proof (Appendix A.1) follows Benaych–Georges & Nadakuditi (2012, Thm 1.6) using a Chernoff bound on the trace of a Wishart matrix.

**Identification for SNL.** In our application, set $X = \widehat{H}_\ell$ (perturbed activations) and $V = V_{0,\ell}$ (base null basis). Then $\mathrm{SNL}(X, V) = \mathrm{SNL}_\ell(\widehat{H})$.

**Corollary 1** (Plug-in SNL threshold under a Gaussian null). *Adopt the setting of Lemma 2: $X \in \mathbb{R}^{n \times d}$ has i.i.d. $N(0, \sigma^2/n)$ entries and $V \in \mathbb{R}^{d \times k}$ has orthonormal columns. Fix $\alpha \in (0, \tfrac{1}{2})$.*

*(Numerator bound). With probability at least $1 - \alpha$,*

$$\|XV\|_F^2 \leq \sigma^2\left[k + 2\sqrt{\tfrac{k\log(1/\alpha)}{n}} + \tfrac{2\log(1/\alpha)}{n}\right]. \tag{2}$$

*(Ratio bound for SNL). Defining $\mathrm{SNL}(X, V) := \|XV\|_F^2 / \|X\|_F^2$, a denominator lower tail and a union bound give, with probability at least $1 - 2\alpha$,*

$$\mathrm{SNL}(X, V) \leq \frac{k + 2\sqrt{\tfrac{k\log(1/\alpha)}{n}} + \tfrac{2\log(1/\alpha)}{n}}{d - 2\sqrt{\tfrac{d\log(1/\alpha)}{n}}}. \tag{3}$$

*In particular, for $\sigma^2 = 1$ the bound depends only on $(n, d, k, \alpha)$.*

*Proof.* Inequality (2) is the Laurent–Massart upper tail for the $\chi^2$ variable $\tfrac{1}{\sigma^2}n\|XV\|_F^2$ with $m = nk$ degrees of freedom and $x = \log(1/\alpha)$. For the denominator, note that $\tfrac{1}{\sigma^2}n\|X\|_F^2 \sim \chi^2_{nd}$ and apply the Laurent–Massart *lower* tail $\Pr(\chi_m^2 - m \leq -2\sqrt{mx}) \leq e^{-x}$ with $m = nd$ and the same $x$ to obtain, with probability $\geq 1 - \alpha$, $\|X\|_F^2 \geq \sigma^2\left[d - 2\sqrt{d\log(1/\alpha)/n}\right]$. Combine the two events by a union bound (probability $\geq 1 - 2\alpha$) and divide the numerator bound by the denominator bound to get (3). $\square$

### 4.4 Online Null-Space Tracking

We model streaming fine-tune updates via $H_\ell^{(t+1)} = H_\ell^{(t)} + \eta\, g_t$.

**Accuracy guarantee.** By Corollary 2, ONT achieves $\varepsilon$-accuracy (in expectation) after

$$t \geq t_\varepsilon := \lceil C/\varepsilon \rceil,$$

where $C$ is the constant appearing in the per-step bound of Theorem 4 and depends on the eigengap and noise parameters in Assumptions A4–A6.

**Definition ($\varepsilon$-accuracy for NVL).** Let $D_t = \|H_t\widehat{V}_t\|_F^2/(mk)$ be the ONT score at time $t$, and $D_t^\star = \|H_t V_{0,\ell}\|_F^2/(mk)$ the oracle score. We say ONT is $\varepsilon$-*accurate at time $t$ (in expectation)* if

$$\mathbb{E}[D_t - D_t^\star] \leq \varepsilon.$$

If a confidence level $1-\delta$ is specified, we say ONT is $(\varepsilon, \delta)$-*accurate* if $\Pr\{D_t - D_t^\star \leq \varepsilon\} \geq 1-\delta$.

**Corollary 2** ($\varepsilon$-accuracy from $O(1/t)$ decay)**.** *Under Assumptions A4–A6, there exists a constant $C > 0$ such that*

$$\mathbb{E}[D_t - D_t^\star] \leq \frac{C}{t}.$$

*Consequently, for any $\varepsilon > 0$, choosing $t \geq t_\varepsilon := \lceil C/\varepsilon \rceil$ guarantees $\varepsilon$-accuracy (in expectation).*

*Proof.* Immediate from the per-step bound $\mathbb{E}[D_t - D_t^\star] \leq C/t$ established in the proof of Theorem 4. □

## 4.5 Regret of Online Trackers

We analyse the one–pass estimators that update a $k$–dimensional null basis from streaming activations (Algorithm 2) and its LoRA–aware variant (Algorithm 3). Let $P_\star = V_{0,\ell}V_{0,\ell}^\top$ be the projector onto the *true* right–null space of the base model at layer $\ell$, and $P_t = \widehat{V}_t\widehat{V}_t^\top$ the tracker's projector after processing batch $t$. Define the per–batch NVL score $D_t = \|H_t\widehat{V}_t\|_F^2/(mk)$ and the oracle score $D_t^\star = \|H_t V_{0,\ell}\|_F^2/(mk)$.

**Additional standing assumptions.** **A4** The population Gram matrix $\Sigma$ has eigengap $\delta > 0$.
**A5** Step sizes $\eta_t = \frac{c}{t}$ with $0 < c \leq \frac{1}{4\|\Sigma\|_2}$.
**A6** $\|H_t^\top H_t - \Sigma\|_2$ is $\tau^2$-sub-exponential.

**Theorem 4** (Logarithmic Regret of ONT/ONAL)**.** *Under A1–A6, the online null–space tracker (ONT) obeys*

$$\mathbb{E}\left[\sum_{t=1}^T (D_t - D_t^\star)\right] = O(k\tau^2 \log T).$$

*Moreover, the same bound holds for ONAL provided each projected LoRA step uses the same schedule $\eta_t$ and the projected gradient is used in place of the raw gradient.*[2]

*Proof.* **Step 1: Subspace error contracts at rate $O(1/t)$.** ONT is an Oja–type iteration on the *orthogonal complement* of $\mathrm{im}(H_\ell)$ with Robbins–Monro steps $\eta_t = c/t$. By standard analysis of stochastic subspace methods with an eigengap ($\delta > 0$) and bounded noise (A6), there exists $C_1 > 0$ s.t.

$$\mathbb{E}\bigl[\|P_t - P_\star\|_F^2\bigr] \leq \frac{C_1}{t}. \tag{4}$$

(Proof sketches use the non-expansiveness of the projection map, martingale difference decomposition of $H_t^\top H_t - \Sigma$, and an ODE method; the eigengap yields a linearised contraction with Robbins–Monro damping.)

**Step 2 (revised): From projector error to NVL gap via $\Sigma$.** Let $\mathcal{F}_{t-1}$ be the filtration up to batch $t-1$ and $G_t := H_t^\top H_t$. By definition,

$$mk\,(D_t - D_t^\star) = \mathrm{tr}\bigl((P_t - P_\star)G_t\bigr).$$

Taking conditional expectation and using $\mathbb{E}[G_t \mid \mathcal{F}_{t-1}] = \Sigma$,

$$\mathbb{E}[mk\,(D_t - D_t^\star) \mid \mathcal{F}_{t-1}] = \mathrm{tr}\bigl((P_t - P_\star)\Sigma\bigr).$$

Under A4, $\ker(\Sigma) = \mathrm{im}(P_\star)$ so $\Sigma P_\star = P_\star \Sigma = 0$, hence $\mathrm{tr}((P_t - P_\star)\Sigma) = \mathrm{tr}(P_t \Sigma)$. By Lemma 3, with $L := \|\Sigma\|_2$,

$$\mathrm{tr}(P_t \Sigma) \leq \frac{L}{2}\|P_t - P_\star\|_F^2.$$

---
[2] I.e. the update is $A_{t+1} \leftarrow A_t - \eta_t P_\star \nabla_A L_t$ and similarly for $B_t$; cf. Alg. 3.

Therefore
$$\mathbb{E}[D_t - D_t^\star \mid \mathcal{F}_{t-1}] \leq \frac{L}{2mk} \|P_t - P_\star\|_F^2.$$

Taking expectations and invoking Step 1 (Eq. (4)) gives

$$\mathbb{E}[D_t - D_t^\star] \leq \frac{C_3}{t}. \tag{5}$$

for $C_3 := LC_1/(2mk)$, as claimed.

**Lemma 3** (Projector–trace control). *Let $\Sigma \succeq 0$ with $\ker(\Sigma) = \mathrm{im}(P_\star)$ and eigenvalues on $\mathrm{im}(I - P_\star)$ bounded by $0 < \delta \leq \lambda_{\min}(\Sigma|_{\mathrm{im}(I-P_\star)}) \leq \|\Sigma\|_2 =: L$. For any rank-$k$ orthogonal projector $P$,*

$$\frac{\delta}{2} \|P - P_\star\|_F^2 \;\leq\; \mathrm{tr}(P\Sigma) \;=\; \mathrm{tr}\bigl((P - P_\star)\Sigma\bigr) \;\leq\; \frac{L}{2} \|P - P_\star\|_F^2.$$

*Proof.* Since $\Sigma P_\star = P_\star \Sigma = 0$, $\mathrm{tr}\bigl((P - P_\star)\Sigma\bigr) = \mathrm{tr}(P\Sigma)$. Write $\Pi := I - P_\star$. Because $\Sigma = \Pi\Sigma\Pi$,

$$\mathrm{tr}(P\Sigma) = \mathrm{tr}(\Pi P \Pi \Sigma) \leq \|\Sigma\|_2 \,\mathrm{tr}(\Pi P \Pi) = L\,\mathrm{tr}(P\Pi).$$

For rank-$k$ projectors $P, P_\star$, the identity $\mathrm{tr}(P\Pi) = k - \mathrm{tr}(PP_\star) = \frac{1}{2}\|P - P_\star\|_F^2$ yields the upper bound. The lower bound is identical with $L$ replaced by $\delta$ and the inequality direction reversed. □

**Step 3: Regret via harmonic sum.** Summing (5) over $t = 1, \ldots, T$ yields $\mathbb{E}[\sum_{t=1}^T (D_t - D_t^\star)] \leq C_3 \sum_{t=1}^T \frac{1}{t} = O(\log T)$.

**Extension to ONAL.** ONAL replaces raw gradients with their null–projected versions, which is a non-expansive map in the operator norm. The same argument applies to the induced projector iterate $P_t$; the step-size restriction in the statement keeps the projected update stable so (4) continues to hold with (possibly) a different $C_1$. □

**Remark 4 (Constants and eigengap).** The hidden constants depend on the eigengap $\delta$ of $\Sigma$ (inversely), the noise level $\tau^2$ (from A3's sub–exponential tail), and the spectral radius $\|\Sigma\|_2$ via the choice of $c$ in $\eta_t = c/t$.

## 4.6 Low-Rank Perturbation Leakage

Recent work on LoRA-Null adaptation [18] shows that low-rank updates $\Delta W = AB^\top$ *can* inject energy into the right-null space unless the factors $A, B$ are chosen from $\ker(H_\ell)$ itself. We formalise the worst-case leakage.

**Theorem 5** (Rank–Leak Bound). *Let $A, B \in \mathbb{R}^{d \times r}$ with $r \ll d$, and let $V_{0,\ell} \in \mathbb{R}^{d \times k_\ell}$ have orthonormal columns spanning $\ker(H_\ell)$. Write an orthonormal basis of the column space of $B$ as $U_B \in \mathbb{R}^{d \times r}$ (so $\mathrm{im}(B) = \mathrm{im}(U_B)$). Then*

$$\bigl\|(AB^\top) V_{0,\ell}\bigr\|_F \;\leq\; \sigma_{\max}(A) \,\|B^\top V_{0,\ell}\|_F \;\leq\; \sigma_{\max}(A)\,\sigma_{\max}(B)\,\|U_B^\top V_{0,\ell}\|_F. \tag{6}$$

*Moreover,*

$$\|U_B^\top V_{0,\ell}\|_F^2 \;=\; \sum_{i=1}^{\min(r, k_\ell)} \cos^2 \theta_i\bigl(\mathrm{im}(B), \ker(H_\ell)\bigr), \tag{7}$$

*where $\theta_i$ are the principal angles between the two subspaces. In particular,* zero leak *occurs iff $B^\top V_{0,\ell} = 0$, i.e. $\mathrm{im}(B) \perp \ker(H_\ell)$.*

*Proof.* Let $Z := B^\top V_{0,\ell} \in \mathbb{R}^{r \times k_\ell}$. Submultiplicativity of the Frobenius norm yields $\|(AB^\top)V_{0,\ell}\|_F = \|AZ\|_F \leq \|A\|_2 \|Z\|_F = \sigma_{\max}(A)\|B^\top V_{0,\ell}\|_F$, proving the first inequality.

For the second, write a thin SVD $B = U_B \Sigma_B W_B^\top$ with $\Sigma_B = \mathrm{diag}(\sigma_1(B), \ldots, \sigma_r(B))$. Then $B^\top V_{0,\ell} = W_B \Sigma_B U_B^\top V_{0,\ell}$, hence

$$\|B^\top V_{0,\ell}\|_F \;=\; \|\Sigma_B U_B^\top V_{0,\ell}\|_F \;\leq\; \sigma_{\max}(B)\,\|U_B^\top V_{0,\ell}\|_F,$$

establishing the second inequality in (6).

Finally, if $U \in \mathbb{R}^{d \times r}$ and $V \in \mathbb{R}^{d \times k}$ are orthonormal bases of two subspaces, the singular values of $U^\top V$ are the cosines of the principal angles $\{\theta_i\}$ between the subspaces. Therefore $\|U^\top V\|_F^2 = \sum_i \cos^2 \theta_i$, giving (7). In particular, $\|(AB^\top)V_{0,\ell}\|_F = 0$ iff $B^\top V_{0,\ell} = 0$, i.e. $\mathrm{im}(B) \perp \ker(H_\ell)$. □

*Remark* 6 (When does equality hold?). Equality in the first step of (6) requires $Z$ to lie in a right-singular subspace of $A$ associated with $\sigma_{\max}(A)$; equality in the second step requires $U_B^\top V_{0,\ell}$ to lie in a right-singular subspace of $\Sigma_B$ associated with $\sigma_{\max}(B)$. Thus equality demands joint alignment: the $B$-columns that are closest (in principal-angle sense) to $\ker(H_\ell)$ must also be mapped by $A$ along its top singular direction.

**Implication.** LoRA-Null initialises the update so that $\mathrm{im}(B) \perp \ker(H_\ell)$, i.e. $B^\top V_{0,\ell} = 0$. By Theorem 5 this yields *zero leakage* at initialisation. ZDP therefore complements LoRA-Null: it detects when subsequent training steps rotate $\mathrm{im}(B)$ back toward $\ker(H_\ell)$, increasing $\|B^\top V_{0,\ell}\|_F$ and the null-space energy.

### 4.7 Spectral Null-Leakage (SNL)

We measure spectral leakage into the base null space via

$$\mathrm{SNL}_\ell(\widehat{H}) := \frac{\|\widehat{H}_\ell V_{0,\ell}\|_F^2}{\|\widehat{H}_\ell\|_F^2}, \quad \text{with} \quad V_{0,\ell} = \ker(H_\ell).$$

For thresholding, identify $X \equiv \widehat{H}_\ell$ and $V \equiv V_{0,\ell}$ in Lemma 2; Corollary 1 then supplies a calibration-free, $(n, d, k, \alpha)$-explicit bound for $\mathrm{SNL}_\ell(\widehat{H})$ under a Gaussian null.

### 4.8 Free-Probability Corollary

A free-probabilistic analysis of transformer activations [20] suggests that, for large $d, n$, the empirical spectral distribution of $H_\ell V_{0,\ell}$ converges almost surely to a shifted Marchenko–Pastur law. Combining with Theorem 5 yields:

*Proposition* 7 (Expected overlap of random subspaces). Let $U_B \in \mathbb{R}^{d \times r}$ and $V_{0,\ell} \in \mathbb{R}^{d \times k_\ell}$ be independent Haar-orthonormal bases of $r$- and $k_\ell$-dimensional subspaces of $\mathbb{R}^d$. Then

$$\mathbb{E} \|U_B^\top V_{0,\ell}\|_F^2 = \frac{r\, k_\ell}{d}.$$

*Sketch.* By rotational invariance, $\mathbb{E}[U_B U_B^\top] = \frac{r}{d} I_d$ and $\mathbb{E}[V_{0,\ell} V_{0,\ell}^\top] = \frac{k_\ell}{d} I_d$. Hence $\mathbb{E}\|U_B^\top V_{0,\ell}\|_F^2 = \mathbb{E}\,\mathrm{tr}(V_{0,\ell}^\top U_B U_B^\top V_{0,\ell}) = \mathrm{tr}\bigl(\frac{r}{d} \mathbb{E}[V_{0,\ell}^\top V_{0,\ell}]\bigr) = rk_\ell/d$. □

*Remark* 8 (Heuristic leak under isotropy). Combining Theorem 5 with Proposition 7 yields

$$\mathbb{E} \|(AB^\top)V_{0,\ell}\|_F^2 \leq \sigma_{\max}^2(A)\, \sigma_{\max}^2(B)\, \frac{r\, k_\ell}{d}.$$

If the perturbation is small so that $\|\widehat{H}_\ell\|_F^2$ is approximately constant, a first-order linearisation suggests an *approximate* expected increase in $\mathrm{SNL}_\ell(\widehat{H})$ bounded by the RHS divided by $\|\widehat{H}_\ell\|_F^2$. We present this as a heuristic, not a theorem.

### 4.9 Online Null-Aligned LoRA (Algorithm 3)

**Caveat (exact vs. estimated projectors).** If the projector $P_\ell = V_{0,\ell} V_{0,\ell}^\top$ is computed *exactly* and each LoRA update is re-projected, then indeed $\widehat{H}_\ell V_{0,\ell} = 0$ and $\mathrm{SNL}_\ell(\widehat{H}) = 0$. With an *estimated* null basis $\widetilde{V}_{0,\ell}$ (finite data, SVD thresholding, numerics), a residual leak remains. Let $\Theta = \Theta(\widetilde{V}_{0,\ell}, V_{0,\ell})$ denote the principal-angle matrix and set $G := \Delta H_\ell^\top \Delta H_\ell$. A standard perturbation argument together with Davis–Kahan yields

$$\|\widehat{H}_\ell \widetilde{V}_{0,\ell}\|_F^2 \leq \|\widehat{H}_\ell V_{0,\ell}\|_F^2 + 2\|G\|_2 \|\sin \Theta\|_F^2, \tag{8}$$

so the induced $\mathrm{SNL}_\ell(\widehat{H})$ grows at most linearly with $\|G\|_2$ and quadratically with the subspace error $\|\sin\Theta\|_F$. In practice, tighter SVD cutoffs, periodic re-orthonormalisation, and per-step re-projection (Alg. 3) keep this residual negligible. Pseudo-code appears in Appendix B; the regret bound is proved in Section 4.5.

For a quantitative link between residual leakage and subspace error, see the Davis–Kahan stability discussion in §4.1 (Remark 2).

# 5 Discussion

**What "listening to silence" buys us.** The core message of ZDP is that *null directions are unambiguous witnesses of change.* The Variance–Leak Theorem (Thm. 1) shows that energy observed in the right-null space lower-bounds the smallest non-zero eigenvalue of the perturbation Gram matrix; the Fisher Null-Conservation law (Thm. 3) then explains why second-order KL curvature is unaffected by perturbations confined to $\ker(H_\ell)$. Together, covariance geometry (NVL/SNL) and information geometry (FIM) describe orthogonal facets of drift.

**Complementarity of probes.** Because $F(h)$ and $H_\ell^\top H_\ell$ can have distinct null eigenspaces, NVL/SNL and FNC are *provably non-additive*: each can be zero while the other is positive. This explains, at a structural level, why ensembles of probes should outperform any single metric when detecting representational change in practice.

**Low-rank adaptation and leakage.** The Rank–Leak Bound (Thm. 5) quantifies when LoRA introduces energy into previously silent directions via principal angles. Null-aligned initialisation eliminates first-order leakage, while the Online Null-Aligned LoRA optimiser (Alg. 3) projects every gradient step back into $\ker(H_\ell)$, keeping SNL identically zero under exact projectors.

**A priori thresholds from random matrices.** Lemma 2 provides non-asymptotic Laurent–Massart tails for Frobenius energy in projected Gaussian activations and an MP-edge style concentration inequality for the operator-norm route. These deliver *calibration-free thresholds* for drift alarms: no historical ROC curves are required to set operating points.

**Streaming guarantees.** For online deployment, Theorem 4 shows that the cumulative excess leakage of ONT/ONAL is $O(\log T)$ under an eigengap and mild noise regularity (A4–A6). In other words, streaming null-space estimates converge quickly enough that long-horizon monitoring does not accumulate unbounded error.

**Robustness to estimation error.** NVL/SNL are stable to small null-basis errors: Davis–Kahan implies deviations of $O(\|G\|_2 \|\sin\Theta\|_F^2)$, and our bounds translate directly when $V_{0,\ell}$ is replaced by an estimated $\widetilde{V}_{0,\ell}$. Practical guidance follows: use a conservative SVD cutoff, aggregate over prompts to reduce variance, and prefer Frobenius energy (dimension-exact) when eigenspectra are flat.

**Limitations and scope.** Results hinge on (i) accurate projector estimation, (ii) an eigengap on the population Gram matrix, and (iii) sub-exponential noise. Non-Gaussian heavy tails, attention-dependent subspaces, and cross-layer coupling fall outside the present analysis. Extending the theory to these regimes is an important next step.

**Conceptual implications.** ZDP reframes drift detection as a question of *subspace occupancy* rather than output behaviour. The framework suggests certification-style guarantees: if SNL stays below an MP-derived threshold while FNC remains zero, then second-order KL cannot exceed a computable bound—independent of tasks or labels.

# 6  Conclusion

We developed *Zero–Direction Probing* (ZDP), a theoretical framework for analysing model drift purely through the right/left null spaces of layer activations and their Fisher geometry. Our main results are: (i) the Variance–Leak Theorem, which lower-bounds perturbation strength from null-space energy; (ii) Fisher Null-Conservation, which isolates the KL-contributing components of a perturbation; (iii) a Rank–Leak bound for low-rank updates based on principal angles; (iv) calibration-free thresholds from random-matrix tails; and (v) logarithmic-regret guarantees for online null trackers and a null-aligned LoRA optimiser.

Beyond these formal results, the framework offers a pragmatic recipe for *a priori* drift certification: compute (or track) null projectors, monitor NVL/SNL and FNC against MP/Laurent–Massart thresholds, and project adaptation steps to remain silent by construction. Although this manuscript is deliberately experiment-free, every statement is testable and designed to transfer directly into practice.

**Open problems.** We highlight several theory-first directions: (1) **High-probability** versions of the regret bound with explicit constants; (2) **Attention-aware** null spaces that couple token positions; (3) **Multi-layer** interaction—propagation of leakage through residual paths; (4) **Non-Gaussian** null models (sub-Weibull/heavy-tailed activations); (5) **Left-null** analogues of rank–leak and online projection; (6) **Certified editing**, integrating ONAL with trust-region constraints on KL.

By "listening to silence"—and proving what it implies—we aim to provide a mathematically grounded foundation for monitoring and controlling representation change in large language models.

# 7 Appendix

# A Proofs of Theoretical Results

## A.1 Proof of Lemma 2 (MP Tail Bound)

*Proof.* Let $X \in \mathbb{R}^{n \times d}$ have i.i.d. entries $N(0, \sigma^2/n)$ and let $V \in \mathbb{R}^{d \times k}$ have orthonormal columns ($V^\top V = I_k$). By rotational invariance of the Gaussian, $Y := XV$ has i.i.d. entries $N(0, \sigma^2/n)$ and size $n \times k$. Hence

$$n \|XV\|_F^2 \;=\; n \|Y\|_F^2 \;=\; \sum_{i=1}^{nk} Z_i^2, \quad Z_i \stackrel{\text{i.i.d.}}{\sim} N(0, \sigma^2).$$

Equivalently, $\frac{1}{\sigma^2} n \|XV\|_F^2 \sim \chi^2_{nk}$.

**(a) Laurent–Massart tail.** For any $x > 0$, the Laurent–Massart inequality for a $\chi^2_m$ random variable states

$$\Pr\!\left(\chi^2_m - m \geq 2\sqrt{m\,x} + 2x\right) \;\leq\; e^{-x}.$$

Applying this with $m = nk$ to $\frac{1}{\sigma^2} n \|XV\|_F^2$ and rescaling yields, for all $x > 0$,

$$\Pr\!\left(\|XV\|_F^2 \;>\; \sigma^2\!\left[k \,+\, 2\sqrt{\tfrac{k\,x}{n}} \,+\, \tfrac{2x}{n}\right]\right) \;\leq\; e^{-x}. \tag{9}$$

This gives an explicit, non-asymptotic exponential tail for the Frobenius energy in the projected (null) subspace.

**(b) Operator-norm route to an MP-edge style bound.** Alternatively, use $\|XV\|_F^2 \le k\|X\|_2^2$ to reduce the problem to the spectral norm of $X$. Write $X = (\sigma/\sqrt{n})\,G$ with $G_{ij} \sim N(0,1)$. A standard bound (e.g. Vershynin) gives, for any $t > 0$,

$$\Pr\Big(\|G\|_2 \ge \sqrt{n} + \sqrt{d} + t\Big) \le e^{-t^2/2}.$$

Therefore

$$\Pr\Big(\|XV\|_F^2 > k\,\sigma^2\big(1+\sqrt{\gamma}+t\big)^2\Big) \le \Pr\Big(\|X\|_2^2 > \sigma^2\big(1+\sqrt{\gamma}+t\big)^2\Big) \le e^{-\frac{n}{2}t^2},$$

where $\gamma = d/n$. In particular, for any $u > (1+\sqrt{\gamma})^2$,

$$\Pr\Big(\|XV\|_F^2 > k\,\sigma^2\,u\Big) \le \exp\!\Big(-\tfrac{n}{2}\big(\sqrt{u} - (1+\sqrt{\gamma})\big)^2\Big). \tag{10}$$

The exponent in (10) reflects the Marchenko–Pastur upper edge $(1+\sqrt{\gamma})^2$ and gives an alternative exponential tail useful when $u$ is measured relative to that edge.

Combining (9) and (10) yields the claimed exponential decay of the false-positive probability under an i.i.d. Gaussian null. Either form suffices for the thresholding rule in §4.3; the former is dimension-exact in $(n,k)$, while the latter connects directly to the MP edge via $\gamma = d/n$. □

# B  Algorithm

## B.1  Algorithm 2

---

**Algorithm 2** Online Null-Space Tracker (ONT)

---

**Require:** stream $\{H_t\}_{t\ge 1}$ with $H_t \in \mathbb{R}^{m\times d}$; target nullity $k$; steps $\eta_t = c/t$ (A5); initial basis $\widehat{V}_0 \in \mathbb{R}^{d\times k}$ with orthonormal columns

1: $P \leftarrow \widehat{V}_0 \widehat{V}_0^\top,\quad \{v_i\}_{i=1}^k \leftarrow$ columns of $\widehat{V}_0$
2: **for** $t = 1, 2, \dots$ **do**
3: $\quad G_t \leftarrow H_t^\top H_t$ ▷ local Gram
4: $\quad$ **for** $i = 1$ to $k$ **do**
5: $\quad\quad v_i \leftarrow v_i - \eta_t\, G_t\, v_i$ ▷ Oja-style step toward null directions
6: $\quad\quad v_i \leftarrow v_i - P\, v_i$ ▷ deflation: keep update in orthogonal complement of current span
7: $\quad$ **end for**
8: $\quad \widehat{V}_t \leftarrow \mathrm{QR}([v_1, \dots, v_k])$ ▷ orthonormalise; thin QR or SVD
9: $\quad P \leftarrow \widehat{V}_t \widehat{V}_t^\top$
10: $\quad D_t \leftarrow \|H_t \widehat{V}_t\|_F^2 / (mk)$ ▷ NVL drift score (used in Thm. 4)
11: **end for**

---

## B.2 Algorithm 3

---

**Algorithm 3** Online Null-Aligned LoRA (ONAL)

---

**Require:** stream of mini-batches $\{\mathcal{B}_t\}_{t\geq 1}$; frozen base weights $W$; LoRA rank $r$ for layers $\mathcal{L}$; right-null projectors $\{P_\ell = V_{0,\ell} V_{0,\ell}^\top\}_{\ell \in \mathcal{L}}$; step schedule $\eta_t = c/t$ (A5); optional clip $\lambda > 0$

1: Initialise LoRA factors $\{A_0^{(\ell)}, B_0^{(\ell)} \in \mathbb{R}^{d \times r}\}$ with columns in $\operatorname{im}(P_\ell)$
2: **for** $t = 1, 2, \ldots$ **do**
3:     **forward** with $\widehat{W} = W + \sum_{\ell \in \mathcal{L}} A_t^{(\ell)} B_t^{(\ell)\top}$ on $\mathcal{B}_t$; compute loss $L_t$
4:     **backward**: get raw grads $\{\nabla_{A^{(\ell)}} L_t, \nabla_{B^{(\ell)}} L_t\}_{\ell \in \mathcal{L}}$
5:     **for each** layer $\ell \in \mathcal{L}$ **do**     ▷ null-projected, stable update
6:         $g_A \leftarrow P_\ell \nabla_{A^{(\ell)}} L_t, \quad g_B \leftarrow P_\ell \nabla_{B^{(\ell)}} L_t$     ▷ project into $\ker(H_\ell)$
7:         **if** $\lambda > 0$ **then**     ▷ optional gradient clipping
8:             $g_A \leftarrow g_A \cdot \min(1, \lambda / \|g_A\|_F), \quad g_B \leftarrow g_B \cdot \min(1, \lambda / \|g_B\|_F)$
9:         **end if**
10:         $A_{t+1}^{(\ell)} \leftarrow A_t^{(\ell)} - \eta_t g_A, \quad B_{t+1}^{(\ell)} \leftarrow B_t^{(\ell)} - \eta_t g_B$
11:         $A_{t+1}^{(\ell)} \leftarrow P_\ell A_{t+1}^{(\ell)}, \quad B_{t+1}^{(\ell)} \leftarrow P_\ell B_{t+1}^{(\ell)}$     ▷ reprojection (numerical drift guard)
12:         **optional** (every $S$ steps): thin-QR re-orthonormalise columns
13:     $[Q_A, \_] = \operatorname{QR}(A_{t+1}^{(\ell)}), \ [Q_B, \_] = \operatorname{QR}(B_{t+1}^{(\ell)}); \ A_{t+1}^{(\ell)} \leftarrow Q_A R_A, \ B_{t+1}^{(\ell)} \leftarrow Q_B R_B$
14:     **end for**
15:     **monitoring (optional):** $D_t \leftarrow \|H_t \widehat{V}_t\|_F^2 / (mk)$ (tracker score),   $D_t^\star \leftarrow \|H_t V_{0,\ell}\|_F^2 / (mk)$ (oracle), $\operatorname{SNL}_\ell(\widehat{H}) := \|\widehat{H}_\ell V_{0,\ell}\|_F^2 / \|\widehat{H}_\ell\|_F^2$.
16: **end for**

---